# CSSDM Ontology to Enable Continuity of Care Data Interoperability

1st Subhashis Das
*BISITE Research Group*
*Dept of Com. Sc., University of Salamanca*
Salamanca, Spain
0000-0001-9663-9009

2nd Debashis Naskar
*BISITE Research Group*
*Dept of Com. Sc., University of Salamanca*
Salamanca, Spain
0000-0003-1980-0756

3rd Sara Rodr´ıguez Gonza´lez
*BISITE Research Group*
*Dept of Com. Sc.,University of Salamanca*
Salamanca, Spain
0000-0002-3081-5177

4th Pamela Hussey
*Faculty of Science & Health*
*Dublin City University*
Dublin, Ireland
0000-0003-2840-9361

*Abstract*—The rapid advancement of digital technologies and recent global pandemic scenarios have led to a growing focus on how these technologies can enhance healthcare service delivery and workflow to address crises. Action plans that consolidate existing digital transformation programs are being reviewed to establish core infrastructure and foundations for sustainable healthcare solutions. Reforming health and social care to personalize home care, for example, can help avoid treatment in overcrowded acute hospital settings and improve the experiences and outcomes for both healthcare professionals and service users. In this information-intensive domain, addressing the interoperability challenge through standards-based roadmaps is crucial for enabling effective connections between health and social care services. This approach facilitates safe and trustworthy data workflows between different healthcare system providers.

In this paper, we present a methodology for extracting, transforming, and loading data through a semi-automated process using a Common Semantic Standardized Data Model (CSSDM) to create personalized healthcare knowledge graph (KG). The CSSDM is grounded in the formal ontology of ISO 13940 ContSys and incorporates FHIR-based specifications to support structural attributes for generating KGs. We propose that the CSSDM facilitates data harmonization and linking, offering an alternative approach to interoperability. This approach promotes a novel form of collaboration between companies developing health information systems and cloud-enabled health services. Consequently, it provides multiple stakeholders with access to high-quality data and information sharing.

## I. INTRODUCTION

The global crisis caused by the ongoing pandemic has significantly altered the functionality of various service-based industries, with health and social care being among the most affected. In response, national health and social care programs are advancing action plans to integrate digital solutions as a foundational element in the future planning of health and social care systems by 2025 National Health Service *(NHS)*. With the rise of digital platforms, there's growing demand for technology-driven, home-based solutions in health and social care. This trend includes smart technology, remote health monitoring, and information sharing with professionals. Effective care plan management requires more than just hospital summaries and Electronic Health Records (EHRs), reflecting the need for integrated digital solutions. It also necessitates considering individual user needs and social settings, such as living conditions, personal history, and the flow of healthcare information from one setting (e.g., primary care) to another (e.g., home care).

From a technology perspective, urgent requirements include a patient-centric approach, defined access and control mechanisms, healthcare data models, and integrated reference architecture models and frameworks. These elements are reported as essential for progressing action plans for scalable digital solutions to support health and social care. This emphasis on digital integration is highlighted, for instance, *21st Century Cures Act: Interoperability, Information Blocking, and the ONC Health IT* by the USA government's Office of the National Coordinator (ONC).

From a standards perspective in Europe, the European Commission prioritizes healthcare interoperability, cross-border treatment, and the involvement of societal stakeholders in the development of Electronic Health Record (EHR) systems. Standards such as *ISO 23903:2021*-Health Informatics—Interoperability and Integration Reference Architecture—Model and Framework emphasize creating ecosystems that offer a harmonized representation to achieve interoperability. These systems are designed to be flexible, scalable, and capable of following a systems-oriented, architecture-centric, ontology-based, and policy-driven approach.

Globally, the International Organization for Standardization (ISO) 13940/DIS–System of Concepts to Support Continuity of Care [1] provides a conceptual framework to address patient needs throughout their care journey. Health Level 7 (HL7) Fast Healthcare Interoperability Resources *(FHIR)* offers detailed specifications for various types of resources used to store data and address queries across a wide range of healthcare-related issues. However, these resources often lack a semantic data model that can integrate data from different care settings.

Recent studies have demonstrated the advantages of using Knowledge Graphs (KGs) for EHR data utilization and providing explicit, explainable results to address healthcare queries over time [2]. KGs represent knowledge, relationships, and data entities within a formal ontological structure, clarifying healthcare concepts contained in the graph.

In this paper, we present a fusion model and subsequent steps to demonstrate how the Web Ontology Language ontology *(OWL) 2* model can enable data integration from different existing legacy systems through a semi-automated mapping process. Our proposed *Common Semantic Standardized Data Model* (CSSDM) aligns with *ISO 23903:2021* Health Informatics—Interoperability and Integration Reference Architecture—Model and Framework.

The CSSDM can align with reference model scenarios such as: an organization handling healthcare and workflow-related data, a public healthcare data controller (e.g., Health Service Executive (HSE)-Ireland, National Services Scotland (NSS), Spanish National Health System (SNS) or electronic Data Research and Innovation Service (eDRIS)), or a service needing to optimize their data preparation pipelines for more efficient use in research experiments. A core issue across these examples is the need to address data heterogeneity and the tedious processes associated with managing it in the daily work of data analysts (e.g., repeatedly solving similar heterogeneity issues to maintain acceptable service quality for clients). Broader related issues include tackling interoperability challenges with external data sources and interfacing with clients unfamiliar with local conventions and practices. The remainder of this paper sets in place research conducted on addressing heterogeneity by describing the related work and current challenges in healthcare data modelling, in Section II. In Section III we present our overall methodology and implementation approach in Section IV, and evaluation in Section V and conclude in Section VI with a discussion on future work plans and an example of a generated knowledge graph from our initial development work.

## II. RELATED WORKS

Socio-technical theories like Actor-Network Theory (ANT) help understand interactions within healthcare networks [3], while the quadruple helix model [4] fosters collaboration among university, industry, government, and public sectors. Ontology-based information models capture complex relations in formal language [5], integrating schemas by applying ontological principles to represent knowledge and interactions within healthcare systems. For instance, relationships such as professional roles (e.g., doctor, patient, nurse), spatial relations (e.g., located-in, address), applied technologies (e.g., mHealth apps, telemedicine), and qualitative performance (e.g., quality of service, drug performance) can be managed using a web ontology language *(OWL)* model. In a complex healthcare system, both social and non-social relations—such as spatial relations, organizational structures, and system interactions—are crucial. Combining ontological principles with social principles can enhance understanding and lead to a more robust socio-technical system (STS) model.

The ontological analysis of complex healthcare networks using ANT on health facilities has been described in a recent study by Iyamu and Mgudlwa [6]. Additionally, a study by the eHealth Research Group from the University of Edinburgh, Scotland highlights the role of ANT in understanding the implementation of information technology developments in healthcare [7]. Although primarily theoretical, this study provided various approaches for dealing with healthcare networks from an ANT perspective.

For instance, the *Yosemite Project* suggested using the Resource Description Framework *(RDF)* to achieve semantic interoperability of all structured healthcare information. However, this study overlooked human involvement in the design process and did not integrate other healthcare schema standards, such as the ISO:13940 System of Concepts to support continuity of care (ContSys). ContSys is essential for connecting different healthcare settings, offering an overarching conceptual model that includes professional healthcare activities, self-care, care by third parties (e.g., family members, personal care assistants, homecare service providers), and all aspects of social care over an individual's life course.

These existing standards, however, do not align with W3C semantic web technologies and linked data [8], which are crucial for creating and maintaining a globally interconnected graph of data. More recently, a paper by Shang et al. (2021) [2] emphasized using knowledge graphs to connect various nonclinical data with electronic health records (EHR) for better decision-making. Knowledge graphs represent knowledge and data entities in a formal ontological structure, making healthcare concepts explicit.

The recently completed H2020 *InteropEHRate* project also demonstrates an interoperability infrastructure using technologies for health data exchange centered on the citizen. However, this project did not implement or align with specific ISO standards, potentially limiting its reusability on an international scale. As part of developing a standards-based roadmap to guide our research, we critiqued several standards to inform our decision-making and development plan. For example, we reviewed Health Informatics-Guidelines for Implementation of *HL7/FHIR based on ISO 13940 and 13606*. However, we noted that neither of these resources has, to date, modified the existing ISO 13940 model nor included any semantic formalism in their initial work. Earlier work we completed, which offered a formal ontology for continuity of care using DOLCE as a top-level ontology and OWL as a formal language, is cited for background reading [9].

Key challenges include the limited involvement of healthcare professionals in designing healthcare information models [10], leading to ICT-driven rather than domain-driven approaches. This results in non-standardized data models that struggle with adoption and interoperability, particularly with FHIR profiles across different regions [11].

Some ongoing projects are attempting to develop a transformation schema to convert FHIR JSON to JSON-LD and then into a Terse RDF Triple Language (Turtle) format. However, these projects are, in our view, not fit for data integration as they do not follow any ontological principles, such as those suggested by the OntoClean methodology [12], nor do they clearly distinguish between structured attributes and classes. Therefore, in this paper, we mainly focus on adopting a collaborative approach to address these gaps. By working with the ISO Health Informatics Technical Committee (ISO/TC 215) and drawing on experience from the EU Horizon 2020 *interopEHRate* project, we provide a summary of the results of our selected methodology in the following section.

## III. METHODOLOGY

We initiate our proposed methodology with two key assumptions. The first assumption is to avoid creating a new ontology or drafting a new standard. Instead, we aim to utilize existing standards to address the healthcare system issues highlighted in the introduction section. The second assumption is reusability. Our rationale for this choice the following considerations RDF/XML was the first RDF format created by W3C and is thus regarded as the standard format. Consequently, most RDF libraries and triplestores output RDF in this format by default. If you want to work with legacy RDF systems or use XML libraries to manipulate your data (since RDF/XML is valid XML), then RDF/XML is the most practical format to use.

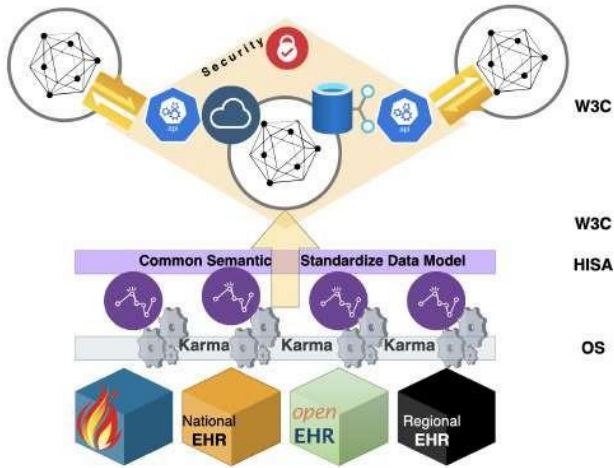

Fig. 1. Transformation process of different datasets using CSSDM, and AI based data integration tool as a middleware; OS Open Source; HISA Health Informatics Service Architecture; W3C World Wide Web Consortium.

Our methodology is based on *ISO/TS 22272:2021*-Health Informatics - Methodology for analysis of business and information needs of health enterprises to support standards-based architectures.

The project's main objective is to create a Common Semantic Standardized Data Model (CSSDM) that can connect old legacy systems using semi-automated mapping and includes an OWL data model to link with all existing Open Linked Data, thereby facilitating and running secondary data analysis. Figure 3, we illustrate the process of mapping data to the CSSDM ontological schema to generate the final ontology-based knowledge graph model. The upper part of the figure, represented in the dark blue box, shows the classes or entity types associated with the ontological schema. The bottom part of the figure displays all the terms in a table format, which we gathered during the processes associated with Extract, transform, and load (ETL). The arrow in the figure represents the direct mapping between data and concepts. This semi-automated process, using the KARMA tool [13] [14], requires supervision by a human expert to ensure accuracy.

We have analyzed and incorporated viewpoints from ISO *ISO 23903:2021* —Interoperability and Integration Reference Architecture, and an inference model for future use in modeling a patient-centric perspective [10]. This groundwork aims to achieve the target state outlined in *ISO/TS 22272:2021*. The overall process of creating the CSSDM is illustrated in Figure 1.

In Step 1, we created a Formal Ontology for Continuity of Care (CSSDM), details of which are available in our previous work [9], [15], [16]. In this step, we considered and consulted existing resources related to information models relevant to continuity of care. These resources included national EHR, regional EHR models, FHIR resources, and Continuity of Care Records (CCR) models, which were then mapped and translated into the formal OWL model.

In Step 2, we presented, discussed, and disseminated information about our formal OWL model to national and international technical committees with which we are engaged. This collaborative effort aimed to agree on and map concepts based on their meanings. For example, we mapped the concept of "subject of care" to *FHIR:Patient*; "observed condition" to *FHIR:Observation* and "Referral" to *FHIR: ServiceRequest*. However, we couldn't find an exact mapping for *FHIR:MedicationRequest* in the Contsys resource, so we created a subclass of "request" instead. The mapping table is provided in the (*supplementary document*).

The expressiveness of our CSSDM ontology model is ALCHQ(D) as per the description logic (DL) scale [17]. We have not exploited the full power of DL-full as supported by the OWL-2 language; rather, we used simple rules to make our model compatible with the GraphDB rule engine and offer quick query execution times.

Finally, in this section, we provide details on the data structure of the class *observation* and the property *person.gender* in the RDF Turtle syntax. The *observation* class reuses properties defined by *FHIR observation resources*. This facilitates our model to be interoperable and semantically aligned with other EHR models using version FHIR 4.6 specification, which is crucial for cross-border studies on intellectual disability clients in the future.

The *observation* captures measurements and makes simple assertions about a *patient*, a device, or another subject. The subject of observation is the Patient class. The performer of the observation is the *healthcare professional* class. A datatype restriction is encapsulated under the property *person.gender*, where data providers must choose among the gender values "Male," "Female," or "Transsexual" as it is mandatory and important information needed by service providers. It cannot be left blank, and the reasoner will detect if incorrect information is inserted into the system. The *supplementary document* demonstrates an example of this observation class detail in RDF Turtle syntax. In Step 3, we provide a summary of the enriched formal OWL model, incorporating attributes specified in the FHIR resources. The figure 2 shows a snapshot from the Protégé editor, highlighting the location of particular concepts in the class hierarchy. For instance, "MedicationRequest" is shown as a subclass of "event". The right side of the figure reflects attributes borrowed from FHIR resources. For instance, *MedicationRequest.note*, *MedicationRequest.status*, *MedicationRequest.dispenseRequest.quantity*

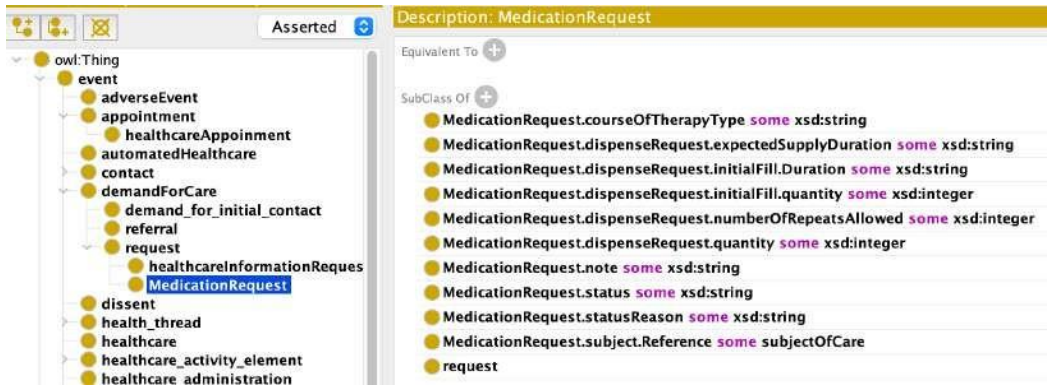

Fig. 2. FHIR:Medication Request attributes inclusion in contSys Request.

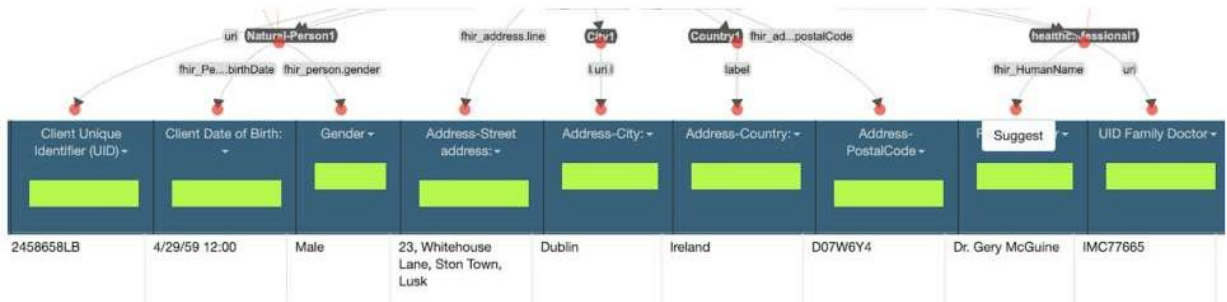

Fig. 3. Semantic Data Mapping using KARMA Data integration tool.

## IV. Technical Implementation

Information on the CSSDM progress is shared with the ISO and CEN Community in phases over two years. For instance, the publication of the formal ontology for continuity of care is accompanied by a supporting blog post on the *Contsys Website*. For the technical implementation, Figure 1 summarizes CSSDM's intended implementation pipeline, detailing the tools and techniques used in our work so far.

We have advanced proposals and funding opportunities for additional data collection. Once ethical approval is obtained, we plan to gather more data from various healthcare sources through our identified service. We expect to achieve this through planned fieldwork, including workshops and surveys, which have proven effective in the preliminary phases of the study.

As the project expanded, the development team cleaned the collected data using *ontoRefine*, a data transformation tool based on OpenRefine integrated into the GraphDB workbench. As the research program grew, larger datasets utilized KARMA for data integration. KARMA, an open-source tool, enabled data integration from various sources such as XML, CSV, text files, and web APIs. Additionally, KARMA generated Relational Databases (RDB) to Resource Description Framework (RDF) Mapping Language *(R2RML)*, facilitating file mapping that could be reused for integrating new or emerging models with new data. Figure 3 depicts the process of converting Irish datasets to RDF data using the CSSDM ontology schema as a common schema. The tool's user interface also helps to check and provide initial mapping by engaging healthcare stakeholders in a loop, thus minimizing disagreement and misalignment. This is a semi-automated process, and we can store the initial mapping file as a R2RML file and reuse it in later processes unless the dataset structure is changed.

For our prototype development, we used the free version of *Ontotext GraphDB* Version 10.6. GraphDB's access control is implemented using a hierarchical role-based access control (RBAC) model, allowing the database administrator to assign specific roles during server setup, such as: **ADMIN**: Can perform all operations without restriction. **USER**: Can save SPARQL queries, graph visualizations, and user-specific settings. **MONITORING**: Allows monitoring operations (queries, updates, abort query/update, resource monitoring). **REPO MANAGER**: Can create, edit, and delete repositories, with read and write permissions to all repositories. GraphDB also supports the lightweight directory access protocol (LDAP).

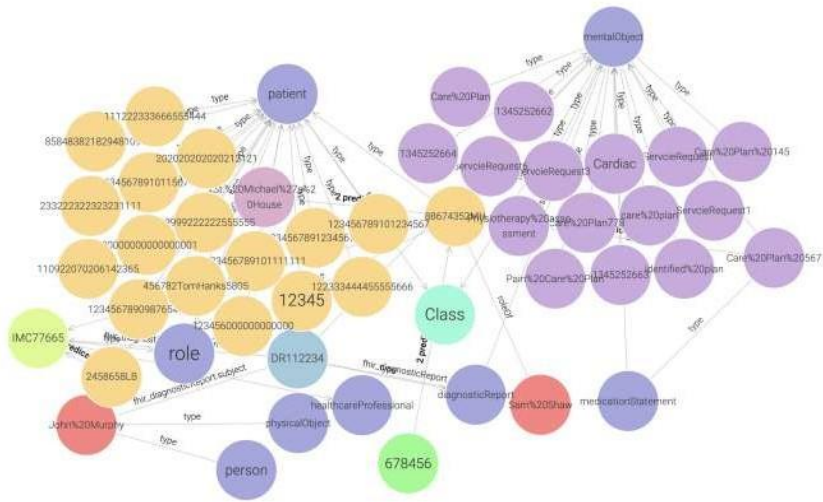

Fig. 4. CSSDM Knowledge Graph depicted specific care plan

This access control and protocol ensure the healthcare system's security by preventing unauthorized third-party access. The CSSDM model can also be deployed on a commercial cloud service provider such as *Amazon Neptune*, where security and safety will be managed by the provider, like Amazon Web Services (AWS).

The practical safety and reliability of CSSDM depend on the healthcare system implementer. In our case, this is handled by *Davra*, an Irish-based startup responsible for large-scale deployment of CSSDM. Davra adheres to several regulatory compliance frameworks, including FedRamp, *HIPAA* compliance, *ISO 27001*, *HITRUST certification*, and *NIST 8259 CSF2014* for cloud software.

## V. Evaluation of the CSSDM Model

This step is highly technical, so the focus of pedagogical instruction at this phase is for domain experts (e.g., healthcare professionals) to supervise and validate the query results produced by data scientists within the healthcare organization. For instance, the main responsibilities of healthcare professionals are to verify and confirm whether the outcomes align with their expectations from the given scenario and their initial thoughts when defining personas.

For example, in a healthcare domain project by the study group, we illustrate some query evaluations. We include several SPARQL queries [18] to demonstrate that the knowledge graph model successfully answered essential competency queries (CQ) [19], potentially useful for further research and analysis. This also showcases how our proposed methodology integrates theory, practice, and interdisciplinarity in a pedagogically effective manner. Consistency of the CSSDM ontology is checked by the *HermiT* reasoner, a Protégé plugin. It identifies subsumption relations between classes. On the other hand Description Logic (DL) based query revealed that model is logically sound in retrieving information. We also have checked CSSDM ontology using *oops!* ontology pitfall scanner and have obtained no pitfall badge.

## VI. Results and Discussion

We define healthcare data interoperability as an ongoing process rather than a one-time task, particularly in the rapidly evolving ICT environment. In this paper, we outline how to take a significant step toward addressing the interoperability challenge by adapting existing models and techniques to fit the Linked Data approach [20], thereby generating an interconnected knowledge graph. Using GraphDB pattern matching, as demonstrated in the following figure 4, helps develop such a graph. Knowledge graphs enable more efficient complex queries compared to typical join operations in relational databases. Figure 5 illustrates the results of CQ1 *Select all female patient who visited hospital and prescribe quantity.* and CQ2 *Select all male patient who visited hospital and prescribe quantity.*, demonstrating how to use Google Chart functionality within the GraphDB SPARQL query panel. It shows that the prescribed quantity of specific drugs is higher for male patients than for female patients. Additional CQs and their SPARQL structures are provided in the *supplementary document* available online.

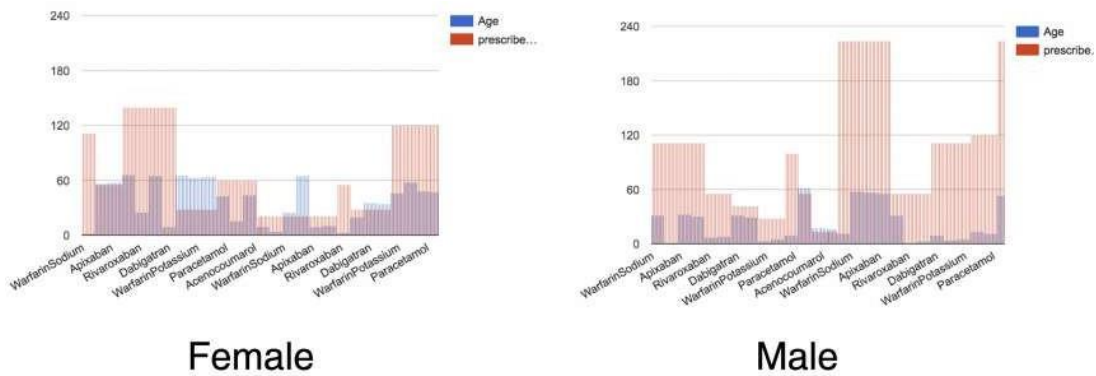

Fig. 5. Comparison chart showing prescribed quantity of Drug X in respect with their age (Result of CQ1, CQ2)

Organizations often struggle with schema-level distinctions, leading to interoperability challenges. It's crucial to differentiate between Common, Core, and Context Schemas [14], mapping only Core and Common attributes while omitting context-specific ones. As a rule of thumb, we recommend that future semantic schema development include only common and core attributes, leaving context-specific attributes to be locally modified based on local needs. This approach advances interoperability for approximately 80% of the data fields. The FHIR-based ContSys semantic schema, available on GitHub ContSysFHIR-SRole.owl.

In phase 1, the CSSDM team identified opportunities for service improvement and plans to showcase the benefits of an interoperable care service architecture, including graph database advantages and data harmonization. Phase 2 will focus on testing user satisfaction using user experience (UX) dimensions from previous work on the Semantic user interface (SemUI) [21]. Our future plan is to test our model with a large dataset one from *MIMIC-IV* free clinical Dataset. The final step will be deploying this system on a cloud server with the help of our industry partner, *Davra*, and conducting performance testing of the CSSDM platform.

ACKNOWLEDGMENT

This project has received funding from the EU Horizon 2020 research and innovation programme under the Marie Skłodowska-Curie grant agreement No. 101034371. The author is indebted to thank his colleagues from *NSAI-HISC, ISO/TC215* , and *CEN/TC251* for their kind and constructive support and cooperation. The project was approved by the President of the Research Ethics of University of Salamanca in Salamanca November, 2023 with document no InformeCEI_EC_en_715.